# The Ability of Image-Language Explainable Models to Resemble Domain Expertise

**Petrus Werner, Anna Zapaishchykova, Ujjwal Ratan**

petruso@kth.se, ujjwalr@amazon.com

## Abstract

Recent advances in vision and language (V+L) models have a promising impact in the healthcare field. However, such models struggle to explain how and why a particular decision was made. In addition, model transparency and involvement of domain expertise are critical success factors for machine learning models to make an entrance into the field. In this work, we study the use of the local surrogate explainability technique to overcome the problem of black-box deep learning models. We explore the feasibility of resembling domain expertise using the local surrogates in combination with an underlying V+L to generate multi-modal visual and language explanations. We demonstrate that such explanations can serve as helpful feedback in guiding model training for data scientists and machine learning engineers in the field.

## 1 Introduction

Studies show that machine learning(ML) models trained on multi-modal inputs outperform models that learn from one modality alone [32], [31], [16], [29]. Inspired by the masked language modeling and next sentence prediction used in the BERT transformer network [12], vision and language (V+L) transformer networks have been created [23], [9] [36]. Several studies in the healthcare domain have confirmed the potential of learning from the joint V+L embedding for clinical tasks, such as the combination of radio-graphs and radiology reports [7], [21], [22]. In [19] authors demonstrate how four multi-modal architectures learn thoracic findings classification task from MIMIC-CXR radiographs and associated reports.

However, even if recent V+L models have a promising impact in the medical field, just like many other SoA deep learning (DL) models, they cannot manifest how and why a model has made a decision. Due to the lack of transparency, such models commonly struggle to enter the medical field, since the model reasoning and explainability are crucial in earning clinicians' trust [10], [28], [4] [30] [8]. One prominent technique to overcome the problem of DL models manifesting as black boxes is to separate the ML model and ML explanations [15]. *Local surrogate explainable models* generate explanations to model predictions by training an inherently interpretative model on model outputs. This explainability technique allows ML practitioners to separate the ML explanations from the model architecture [15].

In addition, as the field of explainable ML has evolved, there is limited work on how to evaluate ML explanations [26] [6]. Yet, in [13], the authors suggest that for application tasks that require extensive human domain expertise, the evaluation of ML explanations should also involve domain expertise. Furthermore, the inclusion of domain expertise has shown to be an important success factor for ML systems to earn trust in the medical field [30] [8]. While there is limited work on multi-modal visual and language explanations, previous work has been carried out on comparing explainability across different types of modality fusion for vision and language learning [2]. However, to the best of our knowledge, there is no previous work done on evaluating visual and language explanations in relation to domain expertise.

This work explores the feasibility of resembling domain expertise using the local surrogate technique to generate multi-modal visual and language explanations. Such explanations provide insights and serve as an interactive feedback loop for learning by highlighting decisive parts of the text and image modality. The study aims to evaluate an ML system as a whole, meaning the combination of an underlying V+L model and the local surrogate explainability technique and its ability to resemble domain expertise.

## 2 Methodology

Firstly, we pre-process the OpenI dataset [11] to extract text tokens and visual features and fine-tune the model on the downstream task of classifying three thoracic findings. We pre-process the text through a BERT encoder network that produces text tokens from the radiology reports [12]. The visual features contain a box position in the X-ray image along with visual embedding vectors. Each image is represented with <36, 4> box position vectors and <36, 2048> visual feature embedding vectors, that were extracted using Detectron2 [37]. The V+L models learn to project the text and visual features to a latent space with the same dimensions [9] [18].

To perform the multi-class classification, a classification head, which generates a probability matrix across the classes of thoracic finding for prediction generation, is added to the joint image and text embedding. We further perform perturbation on the pre-processed tokens and visual features as opposed to raw data [2] to avoid re-extraction of visual features for every new image perturbation [2]. Secondly, by perturbing the direct inputs to the VisualBERT and UNITER model and generating explanations, we create an interactive feedback loop: by perturbing the data and measuring the impact, we find important features that a model seems to have put attention on.

For training, we load pre-trained weights of the VisualBERT and UNITER networks which both have been trained on the COCO data set [20] as well as the VQA 2.0 [1] and Visual Genome [17] data sets. We train using PyTorch inside the Amazon SageMaker ML platform with a GPU accelerated *ml.g4dn.xlarge* instance powered by NVIDIA T4 GPUs.

Dataset  We used the publicly available OpenI dataset [11] for model training and evaluation. The OpenI data set contains 3851 matching data points with radiology reports and chest X-ray images, with labels of 14 thoracic findings from unique patients. The following three labels were selected out of the 14 thoracic findings: Atelectasis, Cardiomegaly, and Nodule.

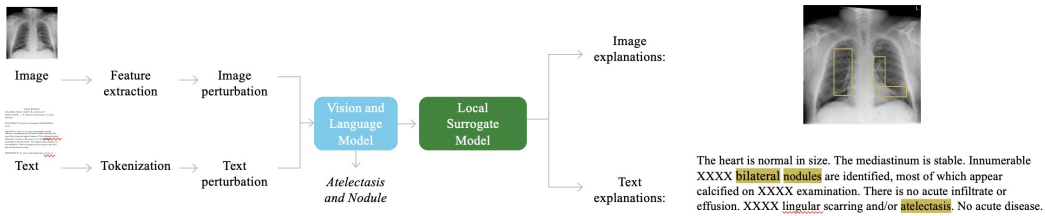

Figure 1: Schematic figure illustrating the training of a local surrogate explainable model.

# 3 Experimental results and discussion

Our goal is to generate multi-modal explanations, i.e., both text and image explanations. To do so, we need to perturb both modalities to fit a simple model that can generate explanations from each modality. We conduct an experiments for two types of explainable models: either we perturb the V+L modality separately and combine the generated explanations afterwards, or we perturb both modalities simultaneously.

| Explainable model: | Approach: | Description: |
|---|---|---|
| Separate Perturbations | Perturb only Text Tokens while keeping Visual Features Fixed + Perturb only Visual Features while keeping Text Tokens Fixed | Finds text and image explanation by combining the outcome of one explainable model that perturbs only the visual embedding and one that perturbs only the tokens. |
| Simultaneous perturbations | Perturb both Modalities Simultaneously | Finds text and image explanations by training an explainable model that perturbs both modalities simultaneously. |

Table 1: Describes the two explainable models of simultaneous and separate perturbations

**Perturb only Text Tokens while keeping Visual Features Fixed** Our implementation is inspired by the python LIME implementation [33]. We add a *wrapper function* for model inference that inputs the perturbed token vectors, while keeping the visual features fixed.

**Perturb only Visual Features while keeping Text Tokens Fixed** The visual features are made from box positions and visual embedding vectors mapped together. We use a binomial distribution function to generate perturbations of the original visual features, where all ones represent the original input vector while the perturbations also contain zeros representing inactivated visual features. The *Probability p* hyperparameter determines the probability of inactivating a feature. The inactivation of a feature means that the associated visual feature vector's elements are set to zeros. This method is inspired by how LIME generates image perturbations, but LIME perturbs pixel regions instead of visual boxes [34].

Furthermore, we assign a weighted score to each perturbation. The *wrapper function* keeps the text tokens fixed this time as opposed to in the case of *Perturb only Text Tokens while keeping Visual Features Fixed* where the visual features were constantly kept the same. Finally, we fit a linear model and, this time, use the coefficients to find the top most decisive perturbations of visual features and collect the inactivated boxes for those perturbations. As an output, the model draws the most decisive boxes onto the input image, which serves as the image explanation.

**Perturb both Modalities Simultaneously** The implementation is a combination of the implementation described in the two previous models, and the inputs are the union of these two.

First of all, the same number of perturbations are generated for the tokens and for the visual features, and weights are computed for each sample of each modality. Moreover, the weights from each modality, derived from the LIME kernel function with values in (0,1) [33], are summed and further normalized to once again fit into the range (0,1). Next, the token and visual perturbations are concatenated, each sample is fed to model inference, and prediction losses are computed. Finally, we fit a linear model and output the top vectors by ranking on their associated coefficients. Similar to in two previously described explainable models, we then output the most decisive words and visual boxes and, together, they serve as a multi-modal explanation.

## 3.1 Evaluation of Explainable Models

Even if the focus of our work is on explainable models, a necessary prerequisite for the case study was to train V+L models on the clinical task of predicting thoracic findings. Figure 2 displays the ROC curves reported on the test set to the UNITER and VisualBERT model across the findings. Both of the models report AUC scores above 0.97 for all three thoracic findings. Yet, the data set size is a severe limitation in our study, and we cannot disclose eventual learned biases which we will further elaborate on in this section.

To evaluate the results of the explainable models, explanations from three radiology domain experts were collected and used as a benchmark [14], [25], [3]. For this, we launched labeling jobs inside the serverless Amazon SageMaker Ground Truth tool [35]. Inside the labeling environment, the domain experts were asked to, given particular thoracic findings, highlight the most explainable set of words of the radiology reports and draw boxes around the most explaining regions of the chest X-ray image.

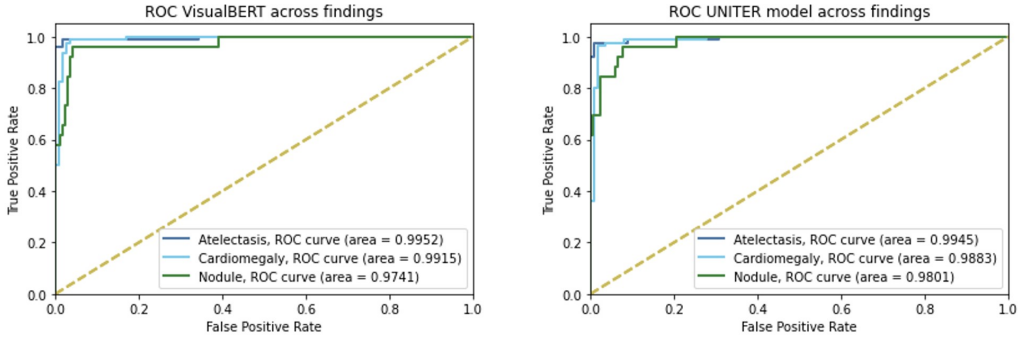

Figure 2: ROC curve for VisualBERT and UNITER models across findings.

Table 2: Displays similarity of text explanations and image explanations between explainable models and domain experts across experiments and types of underlying models. Text explanations refer to identified important words given a prediction of a thoracic finding, and the similarity is computed as the intersection over the union of identified words from the explainable model and domain experts. Image explanations refer to identified important regions of the X-ray images, and the similarity is the intersection of union between the explainable model identified visual boxes and that of domain experts.

| | | Domain expert 1 | | Domain Expert 2 | | Domain Expert 3 | |
|---|---|---|---|---|---|---|---|
| Underlying Model: | Experiment: | Text similarity: | Image similarity: | Text similarity: | Image similarity: | Text similarity | Image similarity |
| UNITER | Simultaneous Perturbations | 0.083 | 0.119 | 0.085 | 0.156 | 0.096 | 0.238 |
| UNITER | Separate Perturbations | 0.103 | 0.102 | 0.122 | 0.172 | 0.138 | 0.261 |
| VisualBERT | Simultaneous Perturbations | 0.073 | 0.091 | 0.079 | 0.016 | 0.100 | 0.261 |
| VisualBERT | Separate Perturbations | 0.128 | 0.102 | 0.171 | 0.172 | 0.117 | 0.302 |
| | Average: | 0.097 | 0.104 | 0.114 | 0.165 | 0.113 | 0.270 |

| | |
|---|---|
| Average text similarity across domain experts: | 0.108 |
| Average image similarity across domain experts: | 0.178 |

As can be seen in Table 2, the similarity scores indicate that the explanations from the explainable model and domain experts are far from identical. The text similarity is rather low, close to 11%, while the image similarity on average is higher, yet only around 18%. However, in discussion with radiology domain experts [14] [25] [3], the experts independently mentioned that reasoning and explanations vary from one domain expert to another for this particular task. To measure the level of variation between annotations gathered from the experts, text and image similarity were computed between each expert. The results can be seen in Table 3. While these comparisons received higher similarity scores compared to those in 2, the scores are still far from an identical score, i.e., a similarity score of 100%. Rather, the similarity scores support notable variations between domain experts' explanations. Table 3: Displays the similarity of text explanations and image explanations between domain experts.

|  | Text Similarity | Image Similarity |
|---|---|---|
| Domain Expert 1 and 2 | 0.62 | 0.38 |
| Domain Expert 1 and 3 | 0.36 | 0.32 |
| Domain Expert 2 and 3 | 0.40 | 0.34 |

Moreover, these results indicate that this is a challenging explanation task and that the ground truth is ambiguous in terms of contributing features from radio-graphs and radiology reports to explain thoracic findings. The similarity scores between the domain experts could represent an upper bound for the explainable model. However, to make further conclusions on our explanations, it could be beneficial to compare with a lower bound. Therefore, instead of generating explanations via our explainable model, we randomly picked words from the radiology reports and visual boxes extracted from the X-ray images to represent random explanations. Then, computing similarity scores for random explanations compared to explanations from domain experts serves as a baseline or a lower bound. Table 4 showcases the similarity scores for the resulting random baseline.

Table 4: Displays the baseline showing text and image similarity domain experts for randomly choosing words and visual boxes.

| Domain Expert | Text Similarity | Image Similarity |
|---|---|---|
| Domain Expert 1 | 0.038 | 0.059 |
| Domain Expert 2 | 0.051 | 0.107 |
| Domain Expert 3 | 0.041 | 0.130 |
| Average | 0.043 | 0.099 |

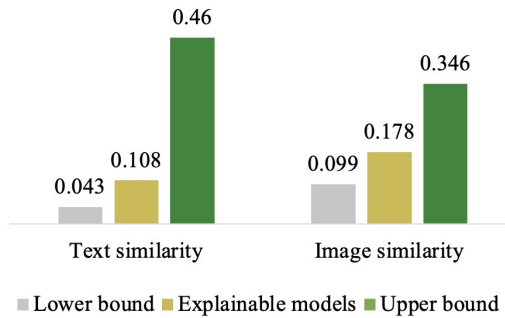

Figure 3: Displays average text and image similarity scores for the lower bound (the random baseline), the explainable models, and the upper bound (the similarity between the domain experts).

Even though the difference between the similarity scores of the random baseline and the explainable model is not massive, one can distinguish a general trend of being better as in Figure 3. Although the generated explanations had relatively low similarity scores to domain experts, the explainable model seems to some extent to have captured signals in explaining the predictions.

Moreover, Figure 4 and 5 showcases three examples each of image explanation and Figure 6 and Figure 7 display three example each of text explanations.

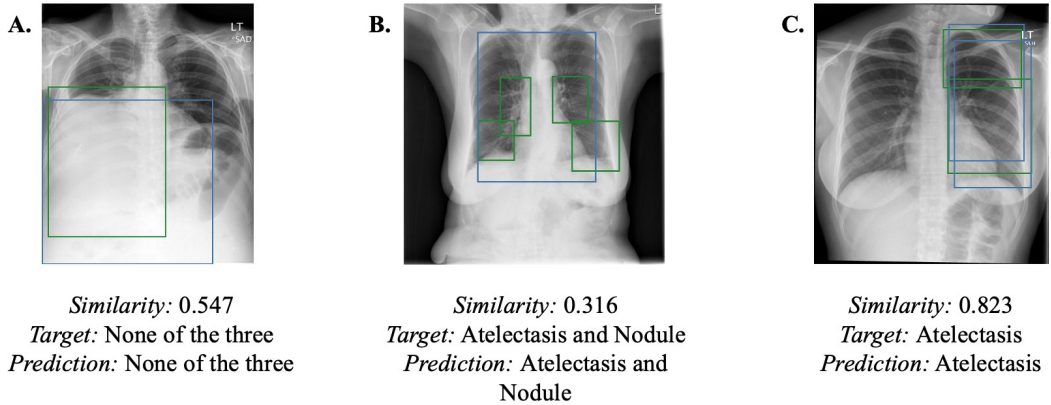

*Similarity:* 0.547  
*Target:* None of the three  
*Prediction:* None of the three

*Similarity:* 0.316  
*Target:* Atelectasis and Nodule  
*Prediction:* Atelectasis and Nodule

*Similarity:* 0.823  
*Target:* Atelectasis  
*Prediction:* Atelectasis

Figure 4: Displays three examples of explanations from explainable models (blue boxes) and explanations from domain expert (green boxes).

One reason for lower image similarity scores is that the set of visual boxes the explainable model could choose from is restricted. When studying Figure 4, it seems that even though the model somewhat matches domain expertise, the boxes look too large and rigid compared to those of the domain experts. Moreover, for example *A* in Figure 5, the underlying UNITER model predicted only atelectasis, but in reality, the ground truth says both atelectasis and cardiomegaly. The domain experts

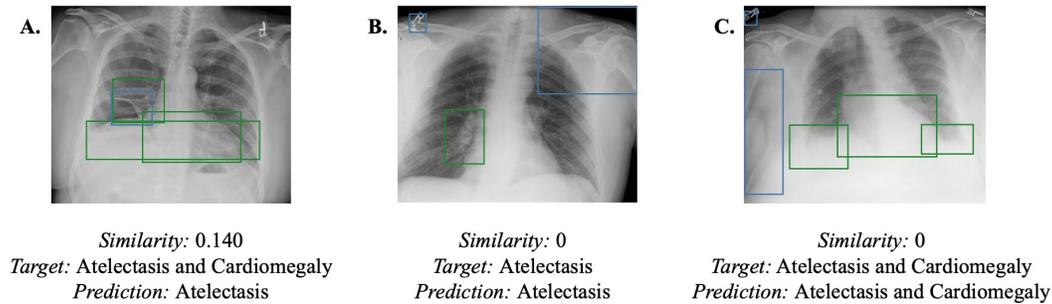

*Similarity:* 0.140  
*Target:* Atelectasis and Cardiomegaly  
*Prediction:* Atelectasis

*Similarity:* 0  
*Target:* Atelectasis  
*Prediction:* Atelectasis

*Similarity:* 0  
*Target:* Atelectasis and Cardiomegaly  
*Prediction:* Atelectasis and Cardiomegaly

Figure 5: Displays three examples of explanations from explainable models (blue boxes) and explanations from domain expert (green boxes).

were given the ground truth and drew their boxes accordingly, represented by the green boxes in the image. As can be seen, the domain expert highlighted the heart region, the horizontal space between the ribs, as well as a region at the mid-left where the latter contains the atelectasis region [25]. Now, looking at the blue box drawn by the explainable model, this one seems to more or less also capture the atelectasis region. However, the explainable model did not highlight the heart region, which typically is associated with cardiomegaly [14]. In this example, it seems that the explanations from the explainable model compared to domain expertise could explain why the model only predicted atelectasis but missed to include cardiomegaly.

Moreover, when studying examples *B* and *C* in Figure 5, one can see that the model seems to focus on the outer parts of the image as well as highlighting letters and numbers in the corner of the images, which generally should be irrelevant for the ML task. Potentially, the model has learned bias in this case. Our data set is relatively small, and we cannot disregard learned biases. One natural way to overcome this is to train the models on a more extensive data set, and it should be a natural next step for future work. Nevertheless, another hypothesis to deal with what looks to be learned biases in detecting the letters and numbers could be to include additional pre-processing steps to the X-ray images. From looking at the explanations provided by the domain experts, we noticed that they consistently are centered around the thorax region. Hence, the outer part of the images seems to be less relevant for predicting the thoracic findings. We hypothesize that by helping the underlying model as well as the explainable model to focus on the thoracic regions of the images, the explainable models could better resemble explanations of explainable models. Future work could explore eventual improvements by, for instance, blurring outer regions of the X-ray images or applying other denoising pre-processing techniques.

**A.** *Similarity:* 0.429  
*Target:* Atelectasis and nodule  
*Prediction:* Atelectasis and nodule  

*Domain Expert:* [innumerable, nodules, atelectasis, bilateral, calcified]  
*Explainable Model:* [innumerable, nodules, atelectasis, or, infiltrate]

**B.** *Similarity:* 0.429  
*Target:* Nodule  
*Prediction:* Nodule  

*Domain Expert:* [calcifications , nodule, process, dense, granulomatous]  
*Explainable Model:* [calcifications , nodule, process, effusion, normal]

**C.** *Similarity:* 0.375  
*Target:* Atelectasis and nodule  
*Prediction:* Atelectasis  

*Domain Expert:* [lung, scarring, calcified, atelectasis, opacities, nodule]  
*Explainable Model:* [lung, scarring, calcified, retrocardiac, effusion]

Figure 6: Displays three examples of words from domain experts and from the explainable models.

For example *A* and *B* in Figure 6 the explainable model manages to capture a majority of the words that the domain expert highlighted, and also the underlying model predicted the correct findings. However, the model predicted atelectasis in example *C* but failed to predict the thoracic finding nodule. Interestingly enough, the explainable words from the explainable model missed to include *nodule* that the domain expert had highlighted in the radiology report

**A.** *Similarity:* 0.1  
*Target:* Atelectasis  
*Prediction:* Atelectasis  

*Domain Expert:* [bibasilar, opacities, atelectasis, costophrenic, blunting, subsegmental]  
*Explainable Model:* [bibasilar, abnormality, costophrenic, lobe, right]

**B.** *Similarity:*0.071  
*Target:* Atelectasis and Cardiomegaly  
*Prediction:* Atelectasis and Cardiomegaly  

*Domain Expert:* [cardiac, silhouette, enlarged, bilateral, opacities, subsegmental, atelectasis, cardiomegaly]  
*Explainable Model:* [cardiac, effusion, right, scattered, pneumothorax]

**C.** *Similarity:* 0.14  
*Target:* None of the three  
*Prediction:* None of the three  

*Domain Expert:* [normal, size, heart]  
*Explainable Model:* [normal, size, appearance, emphysematous, areas]

Figure 7: Displays three examples of words from domain experts and from the explainable models.

One major limitation of the text explanations is that the number of words to output is a pre-defined hyperparameter of the explainable model, which could otherwise improve the results of the similarity scores of the examples displayed in Figure 7. When collecting the language explanations from domain experts, they were allowed to vary the number of words that represent an explanation. In practice, the explainable model always generates a fixed number of words, in this case, five words, while the domain experts' explanations may vary in size from one to eight words. Likely, the fixed size of outputted words harms the text similarity score. Therefore, we considered using the coefficients associated with each of the important vectors to aggregate the importance score of each word and rank the outputted words. Following, one can use this ranking to match the same number of words as the domain experts before computing the similarity. We hypothesize that this would improve our scores but choose not to follow this path. The reason for not doing so is that the information of the size of domain experts' important words would not be available in practice and would serve unfairly in the evaluations. Nevertheless, this is a limitation to our work, and future work is encouraged to study how to fairly vary the size of outputted words of the explainable models. Possibly, introducing a threshold on what words to output could help, but future work needs to explore how to determine such a threshold. In addition, it is worth noting that even the LIME python implementation takes the size of the number of explainable words to output as a pre-defined input parameter [33].

Table 5: Displays average text and image similarity for both underlying vision and language model, as well as for each type of perturbation. The averages derives from the similarity scores presented in 2.

|  | Average Text Similarity | Average Image Similarity |
| --- | --- | --- |
| UNITER | 0.101 | 0.173 |
| VisualBERT | 0.109 | 0.173 |
| Simultaneous Perturbations | 0.088 | 0.122 |
| Separate Perturbations | 0.104 | 0.192 |

Moreover, one interesting take from the experiments presented in Table 2 is the comparison between separate and simultaneous perturbations for training the explainable models. One of the main questions that arose when we designed the explainable model was how to generate the multi-modal explanations. Either one could train separate simple models with their perturbations and then combine the output explanations of the two, or one could train a joint explainable model that trains on perturbations from both modalities. Taking a closer look at the similarity scores in Table 2 and Table 5, one can distinguish that there seems to be a trend that the explainable models with separate perturbations generally receive higher similarity scores than simultaneous perturbation models. For the models with separate perturbations, each surrogate model for the image and text modality could be tuned independently. In contrast, there is an interaction between the hyperparameters from each modality for the explainable model with simultaneous perturbations. We hypothesize that the interaction of hyperparameters of perturbing both modalities simultaneously makes the fine-tuning more complex and challenging to optimize. In addition, it can also be that one of the modalities contains more signal than the other making the dynamics between the simultaneous perturbation more complex, while the annotation from the domain experts does not capture such relationships. The domain experts annotated the modalities separately, which might also favor the separate perturbation technique. For future work, it would be interesting to investigate the strength of the signal from each modality for explaining a prediction.

In contrast, when studying the similarity scores in Table 2 between the UNITER and the VisualBERT model, we could not distinguish any general trend. Instead, the similarity scores from explaining predictions across the two underlying V+L models seem indifferent. Hence, the results suggest that there is no difference between the success of explanations between using an UNITER or a VisualBERT underlying V+L model for this particular case study.

Similarly, another interesting discussion is on ways to perturb features in the input data. Local surrogate explainability techniques like LIME and SHAP build one central assumption: it is possible to turn on and off features for model predictions [34] [24]. While this is an effective way to evaluate feature contributions, it could be worth considering whether such inactivation of features is logical for the ML task. While turning on and off the absence of words and measuring their impact is a common application, there is limited work on how to inactivate the visual boxes. Our approach is to inactivate a visual box by replacing all the elements in the associated visual embedding vector with zeros. As for now, it is difficult to distinguish how this choice impacts the explainable models. Nevertheless, it is worth questioning whether the V+L models are designed to input a visual embedding vector consisting of all zeros. If we were working with pixels, replacing values with zeros would mean that we would get black-colored regions, but for visual embedding, the representation is less intuitive. One other approach could be to compute the mean and standard deviations of values in a visual embedding vector and then distort the elements a multiple of standard deviations away from the mean. Another approach could be to randomize the elements of a visual embedding vector. However, future work is encouraged to explore different ways to represent the inactivation of features for explaining visual and language predictions.

Future work can include using a medical pre-trained visual feature extraction as such a network should help to obtain more relevant features for explanations. For instance, if such a network could detect an area with high opacity, that could be a reasonable explanation for the thoracic finding of atelectasis [14]. However, obtaining such a network is challenging as, to the best of our knowledge, existing medical pre-trained networks only extract embeddings without associated image positions. Such positions, or locations, are necessary to generate visual explanations with our explainable model. However, the development of such a network is challenging due to the lack of data in the field [27].

Overall, comparing the explainable model outcome with domain expertise helps to get insights into whether what the model seems to think are contributing features, do match with what domain expertise view as contributing features. For instance, such insights could suggest to denoise training and inference data for better generalization capabilities or give guidance on hyperparameter-tuning.

## 4 Conclusions & Future Work

This study has explored the feasibility of resembling domain expertise when using the local surrogate explainability technique in combination with an underlying V+L model to generate multi-modal visual and language explanations. A case study was carried out to explain predictions made by V+L models trained to predict thoracic findings from radio-graphs and radiology reports. However, given our experiments, we can not conclude that the explainable models resemble domain expertise. On the other hand, the results indicate that the particular case study task of explaining thoracic findings is challenging as annotations from domain experts suggest that there is ambiguity on what is the ground truth in terms of explanations. Nevertheless, the results indicate that the explainable model has, to some extent, captured signals in explaining the predictions. The resulting similarity scores were relatively far from the similarity levels between domain experts, yet above a random baseline representing the lower bound. In addition, the explainable model captures some useful feedback for model improvement. For instance, explanations could suggest pre-processing data and retraining to better guide the model toward thoracic regions and miss-matches of explanations from the explainable model and domain expertise could potentially serve as an explanation for false negatives. Moreover, the results suggests that the experiments with separate perturbations technique outperform that of simultaneous perturbations in terms of similarity to domain experts. More importantly, the study has identified opportunities for ways to improve the multi-modal V+L explainable model. Future work is encouraged to perform a similar study on another clinical task where there is more unity on the ground truth in terms of explanations. Further, replace the feature extraction network with a medical pre-trained network. Another potential research experiment could include adding a preprocessing step that will segment the area of interest to eliminate the effect of irrelevant signals from the background [5]. It is crucial to explore ways to fairly vary the number of important words and investigate the strength of different modalities when explaining predictions. Finally, study ways to inactivate features and the impact on explanations.

## 5 Acknowledgements